\documentclass[12pt]{article}

\usepackage{scicite}
 \usepackage[superscript]{cite}

\usepackage{times}
\usepackage{amsmath}
\usepackage{graphicx}
\usepackage{lineno}

\newenvironment{sciabstract}{%
\begin{quote} \bf}
{\end{quote}}

\newcounter{lastnote}

\title{A Neural Algorithm of Artistic Style}

\author
{Leon A. Gatys,$^{1,2,3\ast}$ Alexander S. Ecker,$^{1,2,4,5}$ Matthias Bethge$^{1,2,4}$\\
\\
\normalsize{$^{1}$Werner Reichardt Centre for Integrative Neuroscience}\\
\normalsize{and Institute of Theoretical Physics, University of T\"ubingen, Germany}\\
\normalsize{$^{2}$Bernstein Center for Computational Neuroscience, T\"ubingen, Germany}\\
\normalsize{$^{3}$Graduate School for Neural Information Processing, T\"ubingen, Germany}\\
\normalsize{$^{4}$Max Planck Institute for Biological Cybernetics, T\"ubingen, Germany}\\
\normalsize{$^{5}$Department of Neuroscience, Baylor College of Medicine, Houston, TX, USA}
\\
\normalsize{$^\ast$To whom correspondence should be addressed; E-mail:  leon.gatys@bethgelab.org}
}

\date{}

\begin{document} 


\baselineskip24pt


\maketitle

\begin{sciabstract}
In fine art, especially painting, humans have mastered the skill to create unique visual experiences through composing a complex interplay between the content and style of an image. Thus far the algorithmic basis of this process is unknown and there exists no artificial system with similar capabilities. However, in other key areas of visual perception such as object and face recognition near-human performance was recently demonstrated by a class of biologically inspired vision models called Deep Neural Networks \cite{krizhevsky_imagenet_2012, taigman_deepface:_2014}. Here we introduce an artificial system based on a Deep Neural Network that creates artistic images of high perceptual quality. The system uses neural representations to separate and recombine content and style of arbitrary images, providing a neural algorithm for the creation of artistic images. Moreover, in light of the striking similarities between performance-optimised artificial neural networks and biological vision  \cite{guclu_deep_2015,yamins_performance-optimized_2014,cadieu_deep_2014,kummerer_deep_2015,khaligh-razavi_deep_2014}, our work offers a path forward to an algorithmic understanding of how humans create and perceive artistic imagery.\end{sciabstract}

\begin{figure}
\includegraphics[width=1\textwidth]{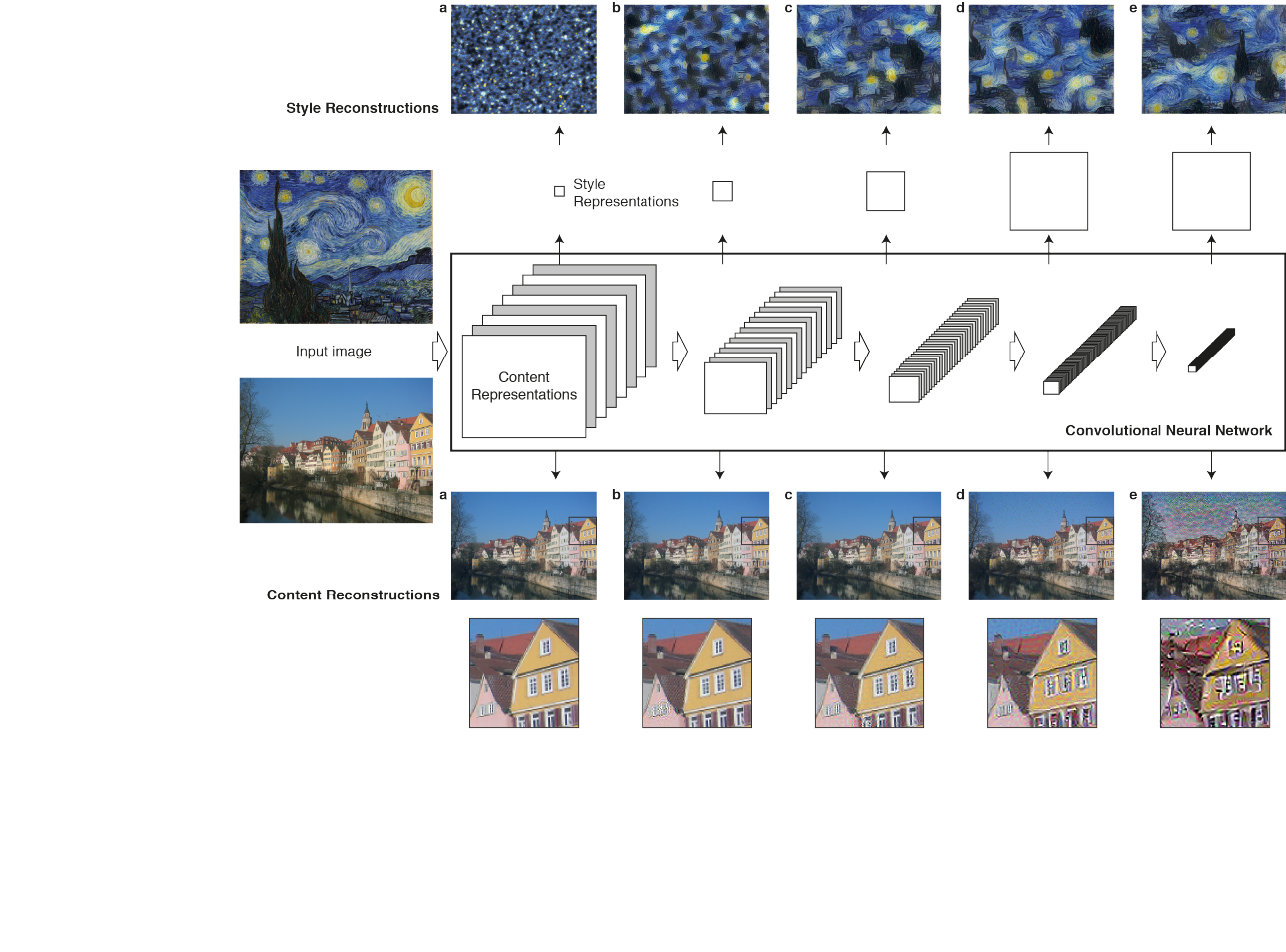}
\caption{\textbf{Convolutional Neural Network (CNN)}. A given input image is represented as a set of filtered images at each processing stage in the CNN. While the number of different filters increases along the processing hierarchy, the size of the filtered images is reduced by some downsampling mechanism (e.g. max-pooling) leading to a decrease in the total number of units per layer of the network. 
\textbf{Content Reconstructions}. We can visualise the information at different processing stages in the CNN by reconstructing the input image from only knowing the network's responses in a particular layer. We reconstruct the input image from from layers `conv1\_1' (\textbf{a}), `conv2\_1' (\textbf{b}), `conv3\_1' (\textbf{c}), `conv4\_1' (\textbf{d}) and `conv5\_1' (\textbf{e}) of the original VGG-Network.
We find that reconstruction from lower layers is almost perfect (\textbf{a},\textbf{b},\textbf{c}). In higher layers of the network, detailed pixel information is lost while the high-level content of the image is preserved (\textbf{d},\textbf{e}).
\textbf{Style Reconstructions}. On top of the original CNN representations we built a new feature space that captures the style of an input image. The style representation computes correlations between the different features in different layers of the CNN. We reconstruct the style of the input image from style representations built on different subsets of CNN layers ( `conv1\_1' (\textbf{a}), `conv1\_1' and `conv2\_1' (\textbf{b}), `conv1\_1', `conv2\_1' and `conv3\_1' (\textbf{c}), `conv1\_1', `conv2\_1', `conv3\_1' and `conv4\_1' (\textbf{d}), `conv1\_1', `conv2\_1', `conv3\_1', `conv4\_1' and `conv5\_1' (\textbf{e})). This creates images that match the style of a given image on an increasing scale while discarding information of the global arrangement of the scene.}\label{network}
\end{figure}

The class of Deep Neural Networks that are most powerful in image processing tasks are called Convolutional Neural Networks. Convolutional Neural Networks consist of layers of small computational units that process visual information hierarchically in a feed-forward manner (Fig \ref{network}). Each layer of units can be understood as a collection of image filters, each of which extracts a certain feature from the input image. Thus, the output of a given layer consists of so-called feature maps: differently filtered versions of the input image. 

When Convolutional Neural Networks are trained on object recognition, they develop a representation of the image that makes object information increasingly explicit along the processing hierarchy \cite{gatys_texture_2015}. Therefore, along the processing hierarchy of the network, the input image is transformed into representations that increasingly care about the actual \emph{content} of the image compared to its detailed pixel values. We can directly visualise the information each layer contains about the input image by reconstructing the image only from the feature maps in that layer \cite{mahendran_understanding_2014} (Fig \ref{network}, content reconstructions, see Methods for details on how to reconstruct the image). 
Higher layers in the network capture the high-level \emph{content} in terms of objects and their arrangement in the input image but do not constrain the exact pixel values of the reconstruction. (Fig \ref{network}, content reconstructions d,e). In contrast, reconstructions from the lower layers simply reproduce the exact pixel values of the original image (Fig \ref{network}, content reconstructions a,b,c). We therefore refer to the feature responses in higher layers of the network as the \textit{content representation}.

To obtain a representation of the \emph{style} of an input image, we use a feature space originally designed to capture texture information \cite{gatys_texture_2015}. This feature space is built on top of the filter responses in each layer of the network. It consists of the correlations between the different filter responses over the spatial extent of the feature maps (see Methods for details). By including the feature correlations of multiple layers, we obtain a stationary, multi-scale representation of the input image, which captures its texture information but not the global arrangement. 

Again, we can visualise the information captured by these style feature spaces built on different layers of the network by constructing an image that matches the style representation of a given input image (Fig \ref{network}, style reconstructions)\cite{heeger_pyramid-based_1995, portilla_parametric_2000}. Indeed reconstructions from the style features produce texturised versions of the input image that capture its general appearance in terms of colour and localised structures. Moreover, the size and complexity of local image structures from the input image increases along the hierarchy, a result that can be explained by the increasing receptive field sizes and feature complexity. We refer to this multi-scale representation as \textit{style representation}.

\begin{figure}
\includegraphics[width=1\textwidth]{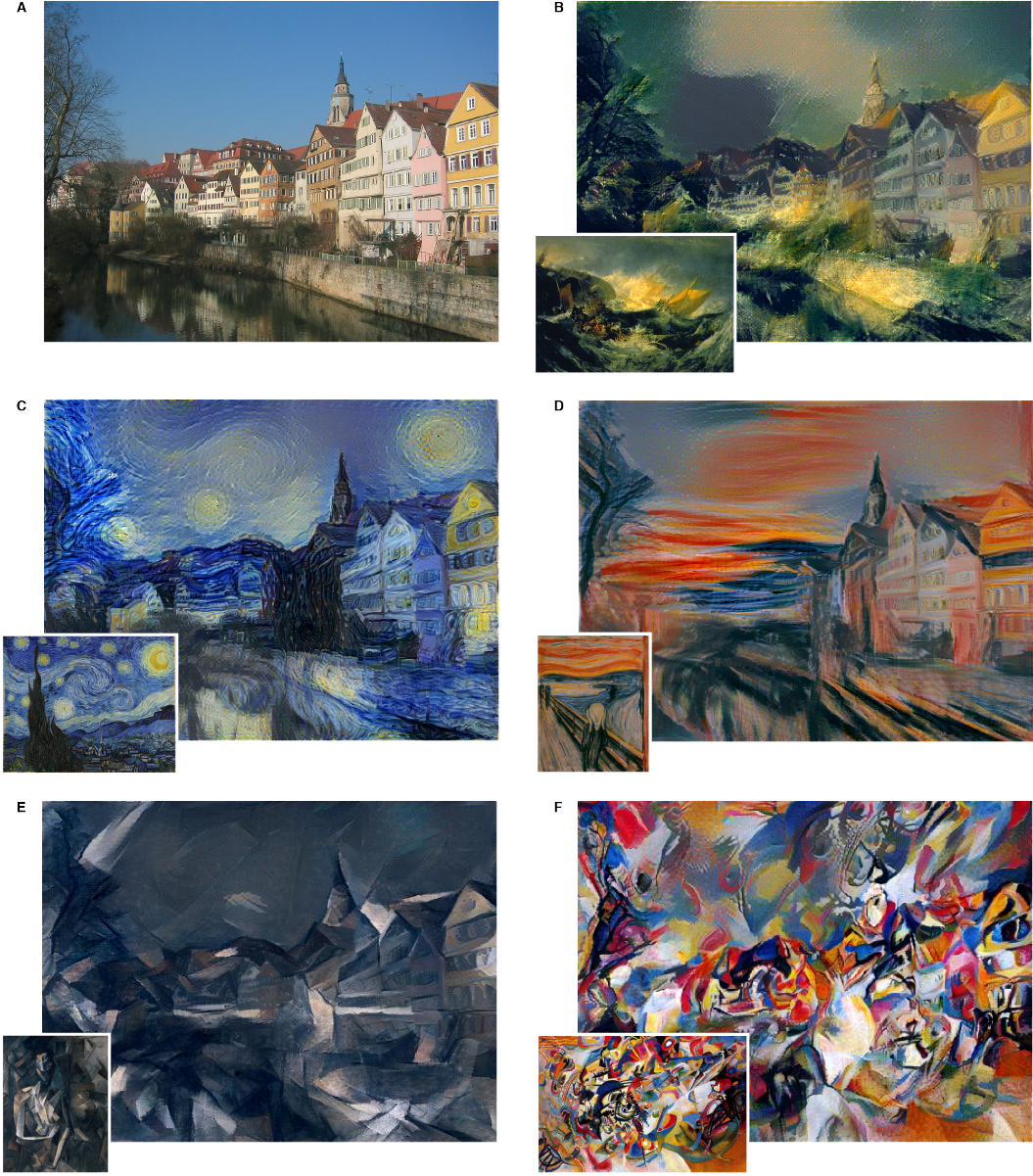}
\caption{Images that combine the content of a photograph with the style of several well-known artworks. The images were created by finding an image that simultaneously matches the content representation of the photograph and the style representation of the artwork (see Methods). The original photograph depicting the Neckarfront in T\"ubingen, Germany, is shown in \textbf{A} (Photo: Andreas Praefcke). The painting that provided the style for the respective generated image is shown in the bottom left corner of each panel.
\textbf{B} \textit{The Shipwreck of the Minotaur} by J.M.W. Turner, 1805. 
\textbf{C} \textit{The Starry Night} by Vincent van Gogh, 1889. 
\textbf{D} \textit{ Der Schrei} by Edvard Munch, 1893. 
\textbf{E}\textit{ Femme nue assise} by Pablo Picasso, 1910. 
\textbf{F}\textit{ Composition VII} by Wassily Kandinsky, 1913.}\label{examples}
\end{figure}

The key finding of this paper is that the representations of content and style in the Convolutional Neural Network are separable. That is, we can manipulate both representations independently to produce new, perceptually meaningful images. To demonstrate this finding, we generate images that mix the content and style representation from two different source images. In particular, we match the content representation of a photograph depicting the ``Neckarfront'' in T\"ubingen, Germany and the style representations of several well-known artworks taken from different periods of art (Fig \ref{examples}).

The images are synthesised by finding an image that simultaneously matches the content representation of the photograph and the style representation of the respective piece of art (see Methods for details). While the global arrangement of the original photograph is preserved, the colours and local structures that compose the global scenery are provided by the artwork. Effectively, this renders the photograph in the style of the artwork, such that the appearance of the synthesised image resembles the work of art, even though it shows the same content as the photograph.

\begin{figure}
\includegraphics[width=1\textwidth]{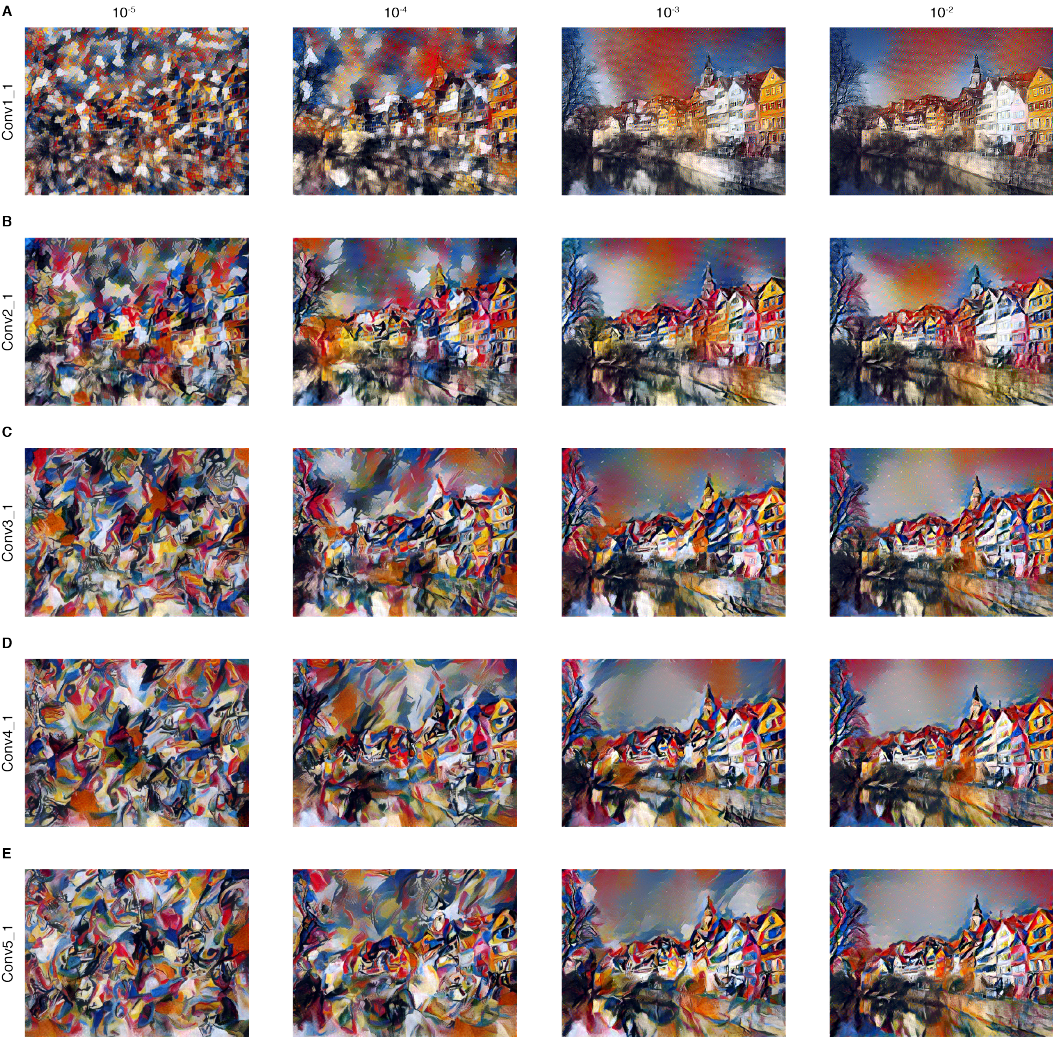}
\caption{Detailed results for the style of the painting \textit{ Composition VII} by Wassily Kandinsky.  The rows show the result of matching the style representation of increasing subsets of the CNN layers (see Methods). We find that the local image structures captured by the style representation increase in size and complexity when including style features from higher layers of the network. This can be explained by the increasing receptive field sizes and feature complexity along the network's processing hierarchy.
The columns show different relative weightings between the content and style reconstruction. The number above each column indicates the ratio $\alpha/\beta$ between the emphasis on matching the content of the photograph and the style of the artwork (see Methods).}\label{detailed}
\end{figure}

As outlined above, the style representation is a multi-scale representation that includes multiple layers of the neural network. In the images we have shown in Fig \ref{examples}, the style representation included layers from the whole network hierarchy. Style can also be defined more locally by including only a smaller number of lower layers, leading to different visual experiences (Fig \ref{detailed}, along the rows). When matching the style representations up to higher layers in the network, local images structures are matched on an increasingly large scale, leading to a smoother and more continuous visual experience. Thus, the visually most appealing images are usually created by matching the style representation up to the highest layers in the network (Fig \ref{detailed}, last row).

Of course, image content and style cannot be completely disentangled. When synthesising an image that combines the content of one image with the style of another, there usually does not exist an image that perfectly matches both constraints at the same time. However, the loss function we minimise during image synthesis contains two terms for content and style respectively, that are well separated (see Methods). We can therefore smoothly regulate the emphasis on either reconstructing the content or the style (Fig \ref{detailed}, along the columns). A strong emphasis on style will result in images that match the appearance of the artwork, effectively giving a texturised version of it, but hardly show any of the photograph's content (Fig \ref{detailed}, first column). When placing strong emphasis on content, one can clearly identify the photograph, but the style of the painting is not as well-matched (Fig \ref{detailed}, last column). For a specific pair of source images one can adjust the trade-off between content and style to create visually appealing images.

Here we present an artificial neural system that achieves a separation of image content from style, thus allowing to recast the content of one image in the style of any other image. We demonstrate this by creating new, artistic images that combine the style of several well-known paintings with the content of an arbitrarily chosen photograph. In particular, we derive the neural representations for the content and style of an image from the feature responses of high-performing Deep Neural Networks trained on object recognition. To our knowledge this is the first demonstration of image features separating content from style in whole natural images. Previous work on separating content from style was evaluated on sensory inputs of much lesser complexity, such as characters in different handwriting or images of faces or small figures in different poses  \cite{tenenbaum_separating_2000,elgammal_separating_2004}. 

In our demonstration, we render a given photograph in the style of a range of well-known artworks. This problem is usually approached in a branch of computer vision called non-photorealistic rendering (for recent review see \cite{kyprianidis_state_2013}). Conceptually most closely related are methods using texture transfer to achieve artistic style transfer \cite{hertzmann_image_2001,ashikhmin_fast_2003, efros_image_2001,lee_directional_2010,xie_feature_2007}. However, these previous approaches mainly rely on non-parametric techniques to directly manipulate the pixel representation of an image. In contrast, by using Deep Neural Networks trained on object recognition, we carry out manipulations in feature spaces that explicitly represent the high level content of an image.

Features from Deep Neural Networks trained on object recognition have been previously used for style recognition in order to classify artworks according to the period in which they were created \cite{karayev_recognizing_2013}. There, classifiers are trained on top of the raw network activations, which we call content representations. We conjecture that a transformation into a stationary feature space such as our style representation might achieve even better performance in style classification.

In general, our method of synthesising images that mix content and style from different sources, provides a new, fascinating tool to study the perception and neural representation of art, style and content-independent image appearance in general. We can design novel stimuli that introduce two independent, perceptually meaningful sources of variation: the appearance and the content of an image. We envision that this will be useful for a wide range of experimental studies concerning visual perception ranging from psychophysics over functional imaging to even electrophysiological neural recordings. In fact, our work offers an algorithmic understanding of how neural representations can independently capture the content of an image and the style in which it is presented. Importantly, the mathematical form of our style representations generates a clear, testable hypothesis about the representation of image appearance down to the single neuron level. The style representations simply compute the correlations between different types of neurons in the network. Extracting correlations between neurons is a biologically plausible computation that is, for example, implemented by so-called complex cells in the primary visual system (V1) \cite{adelson_spatiotemporal_1985}. Our results suggest that performing a complex-cell like computation at different processing stages along the ventral stream would be a possible way to obtain a content-independent representation of the appearance of a visual input. 

All in all it is truly fascinating that a neural system, which is trained to perform one of the core computational tasks of biological vision, automatically learns image representations that allow the separation of image content from style. The explanation could be that when learning object recognition, the network has to become invariant to all image variation that preserves object identity. Representations that factorise the variation in the content of an image and the variation in its appearance would be extremely practical for this task. Thus, our ability to abstract content from style and therefore our ability to create and enjoy art might be primarily a preeminent signature of the powerful inference capabilities of our visual system.

\section*{Methods}
The results presented in the main text were generated on the basis of the VGG-Network \cite{simonyan_very_2014}, a Convolutional Neural Network that rivals human performance on a common visual object recognition benchmark task \cite{russakovsky_imagenet_2014} and was introduced and extensively described in \cite{simonyan_very_2014}. We used the feature space provided by the 16 convolutional and 5 pooling layers of the 19 layer VGG-Network. We do not use any of the fully connected layers.The model is publicly available and can be explored in the caffe-framework \cite{jia_caffe:_2014}. For image synthesis we found that replacing the max-pooling operation by average pooling improves the gradient flow and one obtains slightly more appealing results, which is why the images shown were generated with average pooling.

Generally each layer in the network defines a non-linear filter bank whose complexity increases with the position of the layer in the network.  Hence a given input image $\vec{x}$ is encoded in each layer of the CNN by the filter responses to that image. A layer with $N_l$ distinct filters has $N_l$ feature maps each of size $M_l$, where $M_l$ is the height times the width of the feature map. So the responses in a layer $l$ can be stored in a matrix $F^l \in \mathcal{R}^{N_l \times M_l}$ where $F_{ij}^l$ is the activation of the $i^{th}$ filter at position $j$ in layer $l$. To visualise the image information that is encoded at different layers of the hierarchy (Fig \ref{network}, content reconstructions) we perform gradient descent on a white noise image to find another image that matches the feature responses of the original image. So let $\vec{p}$ and $\vec{x}$ be the original image and the image that is generated and $P^l$ and $F^l$ their respective feature representation in layer $l$. We then define the squared-error loss between the two feature representations
\begin{equation}
\mathcal{L}_{content}(\vec{p},\vec{x},l) = \frac{1}{2}\sum_{i,j}\left(F^l_{ij} - P^l_{ij}\right)^2 \text{ .}
\end{equation}
The derivative of this loss with respect to the activations in layer $l$ equals 
\begin{equation}
\frac{\partial \mathcal{L}_{content}}{\partial F_{ij}^l} =
  \begin{cases}
  \left(F^l - P^l\right)_{ij} & \text{if } F_{ij}^l > 0 \\
   0       & \text{if } F_{ij}^l < 0 \text{ .}
  \end{cases}
\end{equation}
from which the gradient with respect to the image $\vec{x}$ can be computed using standard error back-propagation. Thus we can change the initially random image $\vec{x}$ until it generates the same response in a certain layer of the CNN as the original image $\vec{p}$. The five content reconstructions in Fig \ref{network} are from layers `conv1\_1' (a), `conv2\_1' (b), `conv3\_1' (c), `conv4\_1' (d) and `conv5\_1' (e) of the original VGG-Network.

On top of the CNN responses in each layer of the network we built a style representation that computes the correlations between the different filter responses, where the expectation is taken over the spatial extend of the input image. These feature correlations are given by the Gram matrix $G^l \in \mathcal{R}^{N_l \times N_l}$, where $G_{ij}^l$ is the inner product between the vectorised feature map $i$ and $j$ in layer $l$:
\begin{equation}
G_{ij}^l = \sum_k F_{ik}^l F_{jk}^l.
\label{gram}
\end{equation} 
To generate a texture that matches the style of a given image (Fig \ref{network}, style reconstructions), we use gradient descent from a white noise image to find another image that matches the style representation of the original image. This is done by minimising the mean-squared distance between the entries of the Gram matrix from the original image and the Gram matrix of the image to be generated. 
So let $\vec{a}$ and $\vec{x}$ be the original image and the image that is generated and $A^l$ and $G^l$ their respective style representations in layer $l$. The contribution of that layer to the total loss is then
\begin{equation}
E_l = \frac{1}{4 N_l^2 M_l^2}\sum_{i,j}\left(G^l_{ij}-A^l_{ij}\right)^2
\end{equation}
and the total loss is 
\begin{equation}
\mathcal{L}_{style}(\vec{a},\vec{x}) = \sum_{l=0}^{L}w_{l}E_l
\end{equation}
where $w_l$ are weighting factors of the contribution of each layer to the total loss (see below for specific values of $w_l$ in our results).
The derivative of $E_l$ with respect to the activations in layer l can be computed analytically:
\begin{equation}
\frac{\partial E_l}{\partial F_{ij}^l} =
  \begin{cases}
   \frac{1}{N_l^2 M_l^2}\left((F^l)^{\mathrm T}\left(G^l-A^l\right)\right)_{ji} & \text{if } F_{ij}^l > 0 \\
   0       & \text{if } F_{ij}^l < 0 \text{ .}
  \end{cases}
\end{equation}
The gradients of $E_l$ with respect to the activations in lower layers of the network can be readily computed using standard error back-propagation. The five style reconstructions in Fig \ref{network} were generated by matching the style representations on layer `conv1\_1' (a), `conv1\_1' and `conv2\_1' (b), `conv1\_1', `conv2\_1' and `conv3\_1' (c), `conv1\_1', `conv2\_1', `conv3\_1' and `conv4\_1' (d), `conv1\_1', `conv2\_1', `conv3\_1', `conv4\_1' and `conv5\_1' (e).

To generate the images that mix the content of a photograph with the style of a painting (Fig \ref{examples}) we jointly minimise the distance of a white noise image from the content representation of the photograph in one layer of the network and the style representation of the painting in a number of layers of the CNN. So let $\vec{p}$ be the photograph and $\vec{a}$ be the artwork. The loss function we minimise is

\begin{equation}
\mathcal{L}_{total}(\vec{p},\vec{a},\vec{x}) = \alpha  \mathcal{L}_{content}(\vec{p},\vec{x}) + \beta \mathcal{L}_{style}(\vec{a},\vec{x})
\end{equation}
where $\alpha$ and $\beta$ are the weighting factors for content and style reconstruction respectively. For the images shown in Fig \ref{examples} we matched the content representation on layer `conv4\_2' and the style representations on layers `conv1\_1', `conv2\_1', `conv3\_1', `conv4\_1' and `conv5\_1' ($w_l = 1/5$ in those layers, $w_l = 0$ in all other layers) . The ratio $\alpha/\beta$ was either $1 \times 10^{-3}$ (Fig \ref{examples} B,C,D) or $1 \times 10^{-4}$ (Fig \ref{examples} E,F). Fig \ref{detailed} shows results for different relative weightings of the content and style reconstruction loss (along the columns) and for matching the style representations only on layer `conv1\_1' (A), `conv1\_1' and `conv2\_1' (B), `conv1\_1', `conv2\_1' and `conv3\_1' (C), `conv1\_1', `conv2\_1', `conv3\_1' and `conv4\_1' (D), `conv1\_1', `conv2\_1', `conv3\_1', `conv4\_1' and `conv5\_1' (E). The factor $w_l$ was always equal to one divided by the number of active layers with a non-zero loss-weight $w_l$.

\paragraph{Acknowledgments}
This work was funded by the German National Academic Foundation (L.A.G.), the Bernstein Center for Computational Neuroscience (FKZ 01GQ1002) and the German Excellency Initiative through the Centre for Integrative Neuroscience T\"ubingen (EXC307)(M.B., A.S.E, L.A.G.)
\bibliography{NeuralArt}
\bibliographystyle{naturemag}

\end{document}